%% file: scenic-cav.tex
\documentclass[runningheads]{llncs}
\usepackage{amsmath,amssymb}
\usepackage{fancyvrb}
\usepackage{times}
\usepackage{wrapfig}
\usepackage{graphicx}
\usepackage{xcolor}

\usepackage{hyperref}

\graphicspath{{pics/}}

\input{macros}

\begin{document}
\title{Formal Analysis and Redesign of a Neural Network-Based Aircraft Taxiing System with \verifai{}}
\titlerunning{Formal Analysis and Redesign of an Aircraft Taxiing System with \verifai{}}

\author{Daniel J. Fremont\inst{1,2} \and
Johnathan Chiu\inst{2} \and\\
Dragos D. Margineantu\inst{3} \and
Denis Osipychev\inst{3} \and
Sanjit A. Seshia\inst{2}}
\authorrunning{D. J. Fremont et al.}  
%
\institute{University of California, Santa Cruz, USA \and
University of California, Berkeley, USA \and
Boeing Research \& Technology, Seattle, USA}
\maketitle              
\begin{abstract}
We demonstrate a unified approach to rigorous design of safety-critical autonomous systems using the \verifai{} toolkit for formal analysis of AI-based systems.
\verifai{} provides an integrated toolchain for tasks spanning the design process, including modeling, falsification, debugging, and ML component retraining.
We evaluate all of these applications in an industrial case study on an experimental autonomous aircraft taxiing system developed by Boeing, which uses a neural network to track the centerline of a runway.
We define runway scenarios using the \textsc{Scenic} probabilistic programming language, and use them to drive tests in the X-Plane flight simulator.
We first perform falsification, automatically finding environment conditions causing the system to violate its specification by deviating significantly from the centerline (or even leaving the runway entirely).
Next, we use counterexample analysis to identify distinct failure cases, and confirm their root causes with specialized testing.
Finally, we use the results of falsification and debugging to retrain the network, eliminating several failure cases and improving the overall performance of the closed-loop system.

\end{abstract}
\input{introduction.tex}

\input{approach.tex}

\input{experiments.tex}

\input{conclusion.tex}

\paragraph{\textbf{Acknowledgments.}}
The authors are grateful to Forrest Laine and Tyler Staudinger for assistance with the experiments and TaxiNet, to Ankush Desai for suggesting using \scenic{} as a prior for cross-entropy sampling, and to the anonymous reviewers.

This work was supported in part by NSF grants 1545126 (VeHICaL), 1646208, 1739816, and 1837132, the DARPA BRASS (FA8750-16-C0043) and Assured Autonomy programs, Toyota under the iCyPhy center, and Berkeley Deep Drive.

\bibliographystyle{splncs04}
\bibliography{biblio}

\appendix

\input{more-experiments}

\end{document}

%% file: macros.tex
\newcommand{\scenic}{\textsc{Scenic}}
\newcommand{\verifai}{\textsc{VerifAI}}

\newcommand{\taxinet}{TaxiNet}

\newcommand{\sfalsif}{\ensuremath{\mathcal{S}_{\text{falsif}}}}

\newcommand{\tngen}{\ensuremath{\mathcal{T}_{\text{generic}}}}
\newcommand{\tnhalton}{\ensuremath{\mathcal{T}_{\text{Halton}}}}
\newcommand{\tnspec}{\ensuremath{\mathcal{T}_{\text{specialized}}}}

\newcommand{\alwaysspec}{\ensuremath{\varphi_{\text{always}}}}
\newcommand{\evenspec}{\ensuremath{\varphi_{\text{eventually}}}}

\DefineVerbatimEnvironment{scenario}{Verbatim}{samepage=true,fontfamily=cmtt}
\DefineVerbatimEnvironment{bscenario}{BVerbatim}{samepage=true,fontfamily=cmtt}

%% file: introduction.tex
\section{Introduction} \label{sec:intro}

The expanding use of machine learning (ML) in safety-critical applications has led to an urgent need for rigorous design methodologies that can ensure the reliability of systems with ML components~\cite{russell-letter-2,seshia-arxiv16}.
Such a methodology would need to provide tools for \emph{modeling} the system, its requirements, and its environment, \emph{analyzing} a design to find failure cases, \emph{debugging} such cases, and finally \emph{synthesizing} improved designs.

\begin{wrapfigure}[12]{r}{0.60\textwidth}
    \centering
    \includegraphics[width=0.53\textwidth]{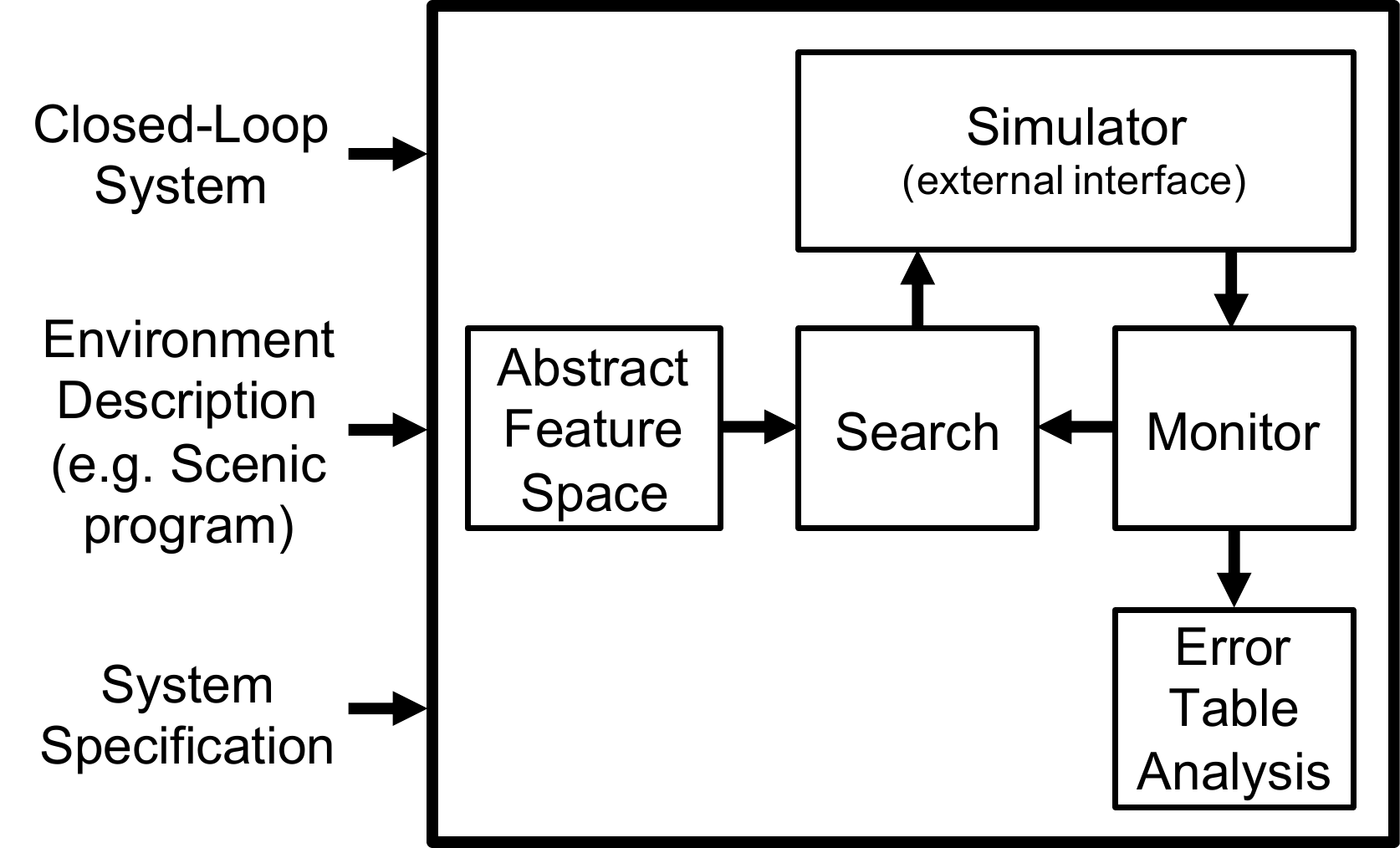}
    \caption{Architecture of \verifai{}.}
    \label{fig:verifai}
\end{wrapfigure}

The \verifai{} toolkit~\cite{verifai} provides a unified framework for all of these design tasks, based on a simple paradigm: simulation driven by formal models and specifications.
The top-level architecture of \verifai{} is shown in Fig.~\ref{fig:verifai}.
In \verifai{}, we first parametrize the space of environments and system configurations of interest, either by explicitly defining parameter ranges or using the \scenic{} probabilistic environment modeling language~\cite{scenic}.
\verifai{} then generates concrete tests by searching this space, using a variety of algorithms ranging from simple random sampling to global optimization techniques.
Finally, we simulate the system for each test, monitoring the satisfaction or violation of a system-level specification; the results of each test are used to guide further
\pagebreak
\noindent
search, and any violations are recorded in a table for automated analysis (e.g. clustering) or visualization.
This architecture enables a wide range of use cases, including falsification, fuzz testing, debugging, data augmentation, and parameter synthesis; Dreossi et al.~\cite{verifai} demonstrated all of these applications individually through several small case studies.

In this paper, we provide an \emph{integrated} case study, applying \verifai{} to a complete design flow for a large, realistic system from industry: \taxinet{}, an experimental autonomous aircraft taxiing system developed by Boeing for the DARPA Assured Autonomy project.
This system uses a neural network to estimate the aircraft's position from a camera image; a controller then steers the plane to track the centerline of the runway.
The main requirement for \taxinet{}, provided by Boeing, is that it keep the plane within 1.5~m of the centerline; we formalized this as a specification in Metric Temporal Logic (MTL)~\cite{mtl}.
Verifying this specification is difficult, as the neural network must be able to handle the wide range of images resulting from different lighting conditions, changes in runway geometry, and other disturbances such as tire marks on the runway.

Our case study illustrates a complete iteration of the design flow for \taxinet{}, analyzing and debugging an existing version of the system to inform an improved design.
Specifically, we demonstrate:
\begin{enumerate}
\item Modeling the environment of the aircraft using the \scenic{} language.
\item Falsifying an initial version of \taxinet{}, finding environment conditions under which the aircraft significantly deviates from the centerline.
\item Analyzing counterexamples to identify distinct failure cases and diagnose potential root causes.
\item Testing the system in a targeted way to confirm these root causes.
\item Designing a new version of the system by retraining the neural network based on the results of falsification and debugging.
\item Validating that the new system eliminates some of the failure cases in the original system and has higher overall performance.
\end{enumerate}

Following the procedure above, we were able to find several scenarios where \taxinet{} exhibited unsafe behavior.
For example, we found the system could not properly handle intersections between runways.
More interestingly, we found that \taxinet{} could get confused when the shadow of the plane was visible, which only occurred during certain times of day and weather conditions.
We stress that these types of failure cases are meaningful counterexamples that could easily arise in the real world, unlike pixel-level adversarial examples~\cite{goodfellowSS14}; we are able to find such cases because \verifai{} searches through a space of \emph{semantic} parameters~\cite{dreossi-cav18}.
Furthermore, these counterexamples are \emph{system-level}, demonstrating undesired behavior from the complete system rather than simply its ML component.
Finally, our work differs from other works on validation of cyber-physical systems with ML components (e.g.~\cite{deeptest}) in that we address a broader range of design tasks (including debugging and retraining as well as testing) and also allow designers to \emph{guide} search by encoding domain knowledge using \scenic{}.

For our case study, we extend \verifai{} in two ways.
First, we interface the toolkit to the X-Plane flight simulator~\cite{xplane} in order to run closed-loop simulations of the entire system, with X-Plane rendering the camera images and simulating the aircraft dynamics.
More importantly, we extend the \scenic{} language to allow it to be used in combination with \verifai's active sampling techniques.
Previously, as in any probabilistic programming language, a \scenic{} program defined a fixed distribution~\cite{scenic}; while adequate for modeling particular scenarios, this is incompatible with active sampling, where we change how tests are generated over time in response to feedback from earlier tests.
To reconcile these two approaches, we extend \scenic{} with \emph{parameters} that are assigned by an external sampler.
This allows us to continue to use \scenic{}'s convenient syntax for modeling, while now being able to use not only random sampling but optimization or other algorithms to search the parameter space.

Adding parameters to \scenic{} enables important new applications.
For example, in the design flow we described above, after finding through testing some rare event which causes a failure, we need to generate a dataset of such failures in order to retrain the ML component.
Na\"ively, we would have to manually write a new \scenic{} program whose distribution was concentrated on these rare events (as was done in~\cite{scenic}).
With parameters, we can simply take the generic \scenic{} program we used for the initial testing, and use \verifai's cross-entropy sampler~\cite{verifai,cross-entropy} to automatically converge to such a distribution~\cite{DBLP:conf/hybrid/SankaranarayananF12}.
Alternatively, if we have an intuition about where a failure case may lie, we can use \scenic{} to encode this domain knowledge as a \emph{prior} for cross-entropy sampling, helping the latter to find failures more quickly.

In summary, the novel contributions of this paper are:
\begin{itemize}
\item The first demonstration on an industrial case study of an integrated toolchain for falsification, debugging, and retraining of ML-based autonomous systems.

\item An interface between \verifai{} and the X-Plane flight simulator.

\item An extension of the \scenic{} language with parameters, and a demonstration using it in conjunction with cross-entropy sampling to learn a \scenic{} program encoding the distribution of failure cases.
\end{itemize}

We begin in Sec.~\ref{sec:approach} with a discussion of our extension of \scenic{} with parameters and our X-Plane interface.
Section~\ref{sec:experiments} presents the experimental setup and results of our case study, and we close in Sec.~\ref{sec:conclusion} with some conclusions and directions for future work.

%% file: approach.tex
\section{Extensions of \verifai{}} \label{sec:approach}

\paragraph{\textbf{\scenic{} with Parameters.}}

To enable search algorithms other than random sampling to be used with \scenic{} we extend the language with a concept of \emph{external parameters} assigned by an \emph{external sampler}.
A \scenic{} program can specify an external sampler to use; this sampler will define the allowed types of parameters, which can then be used in the program in place of any distribution.
The default external sampler provides access to the \verifai{} samplers and defines parameter types corresponding to \verifai{}'s continuous and discrete ranges.
Thus for example one could write a \scenic{} program which picks the colors of two cars randomly according to some realistic distribution, but chooses the distance between them using \verifai{}'s Bayesian Optimization sampler.

The semantics of external parameters is simple: when sampling from a \scenic{} program, the external sampler is first queried to provide values for all the parameters; the program is then equivalent to one without parameters, and can be sampled as usual\footnote{One complication arises because \scenic{} uses rejection sampling to enforce constraints: if a sample is rejected, what value should be returned to active samplers that expect feedback, e.g.~a cross-entropy sampler? By default we return a special value indicating a rejection occurred.}.

\paragraph{\textbf{X-Plane Interface.}}

Our interface between X-Plane and \verifai{} uses the latter's client-server architecture for communicating with simulators.
The server runs inside \verifai{}, taking each generated feature vector and sending it to the client.
The client runs inside X-Plane and calls its APIs to set up and execute the test, reporting back information needed to monitor the specifications.
For our client, we used X-Plane Connect~\cite{xpc}, an X-Plane plugin providing access to X-Plane's ``datarefs''.
These are named values which represent simulator state, e.g.~positions of aircraft and weather conditions.
Our interface exposes all datarefs to \scenic{}, allowing arbitrary distributions to be placed on them.
We also set up the \scenic{} coordinate system to be aligned with the runway, performing the appropriate conversions to set the raw position datarefs.

%% file: experiments.tex
\section{\taxinet{} Case Study} \label{sec:experiments}

\subsection{Experimental Setup} \label{sec:setup}

\begin{figure}[tb]
\centering
\includegraphics[width=0.49\textwidth]{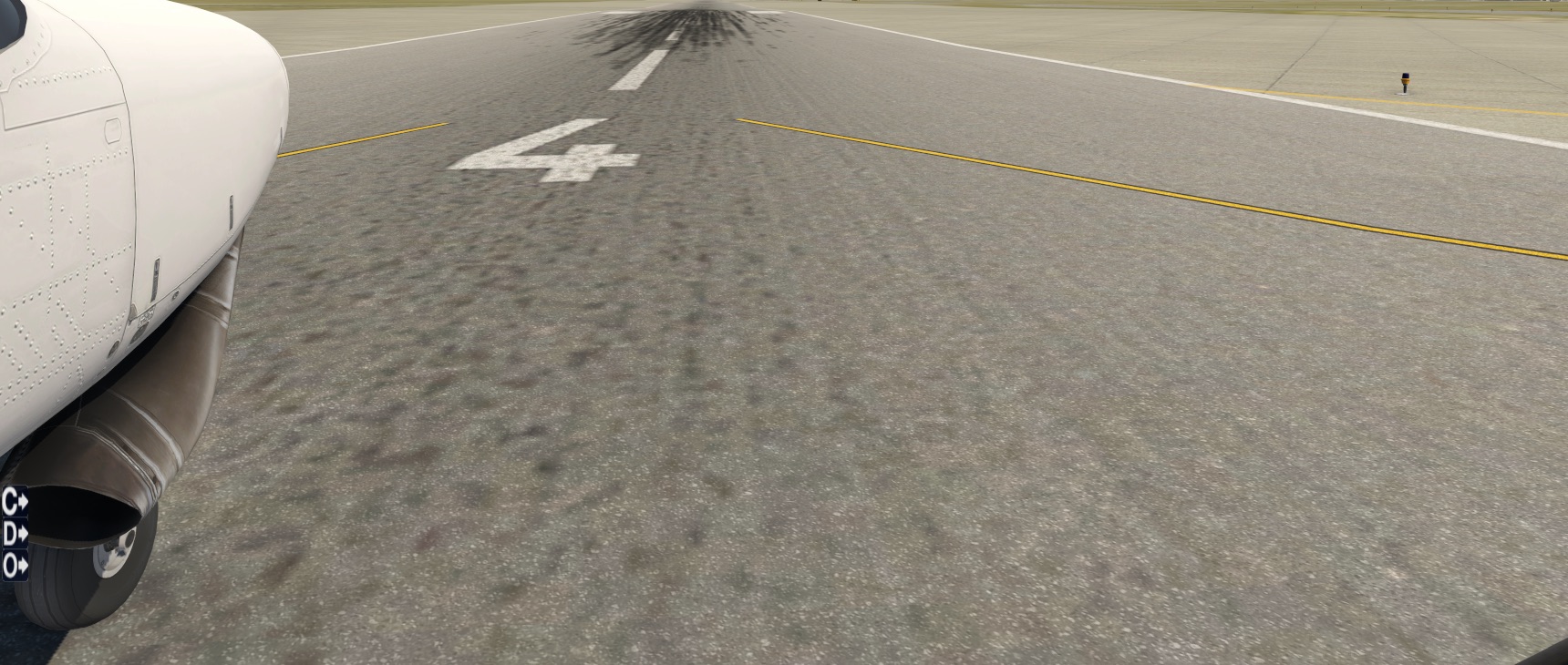}
\includegraphics[width=0.49\textwidth]{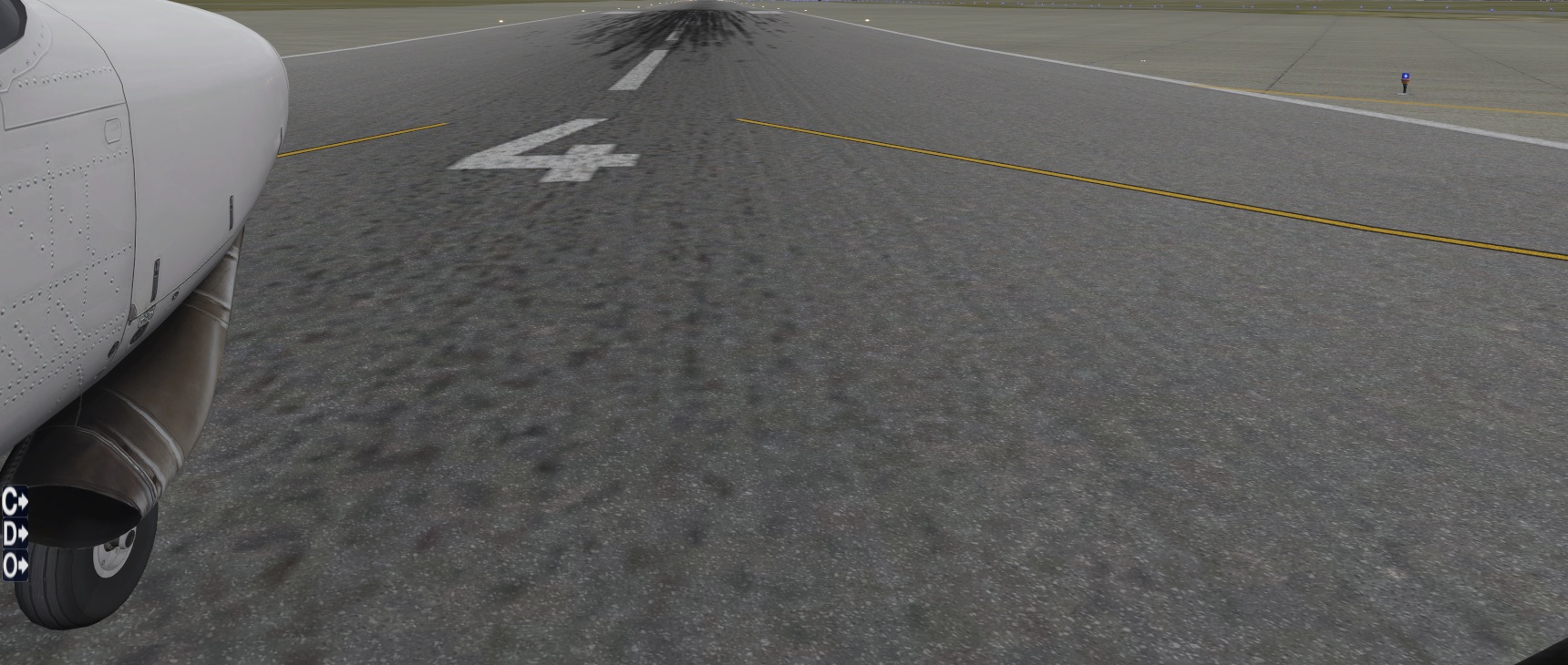}
\includegraphics[width=0.49\textwidth]{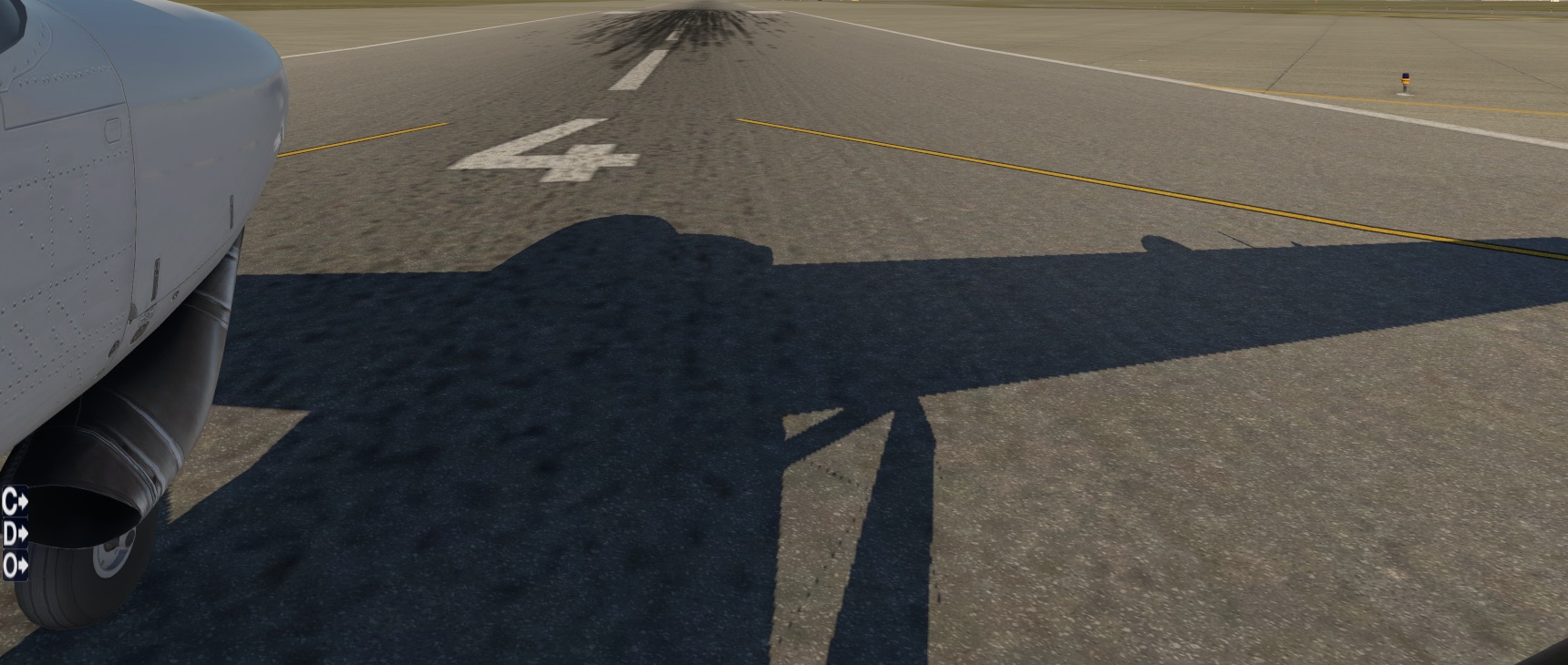}
\includegraphics[width=0.49\textwidth]{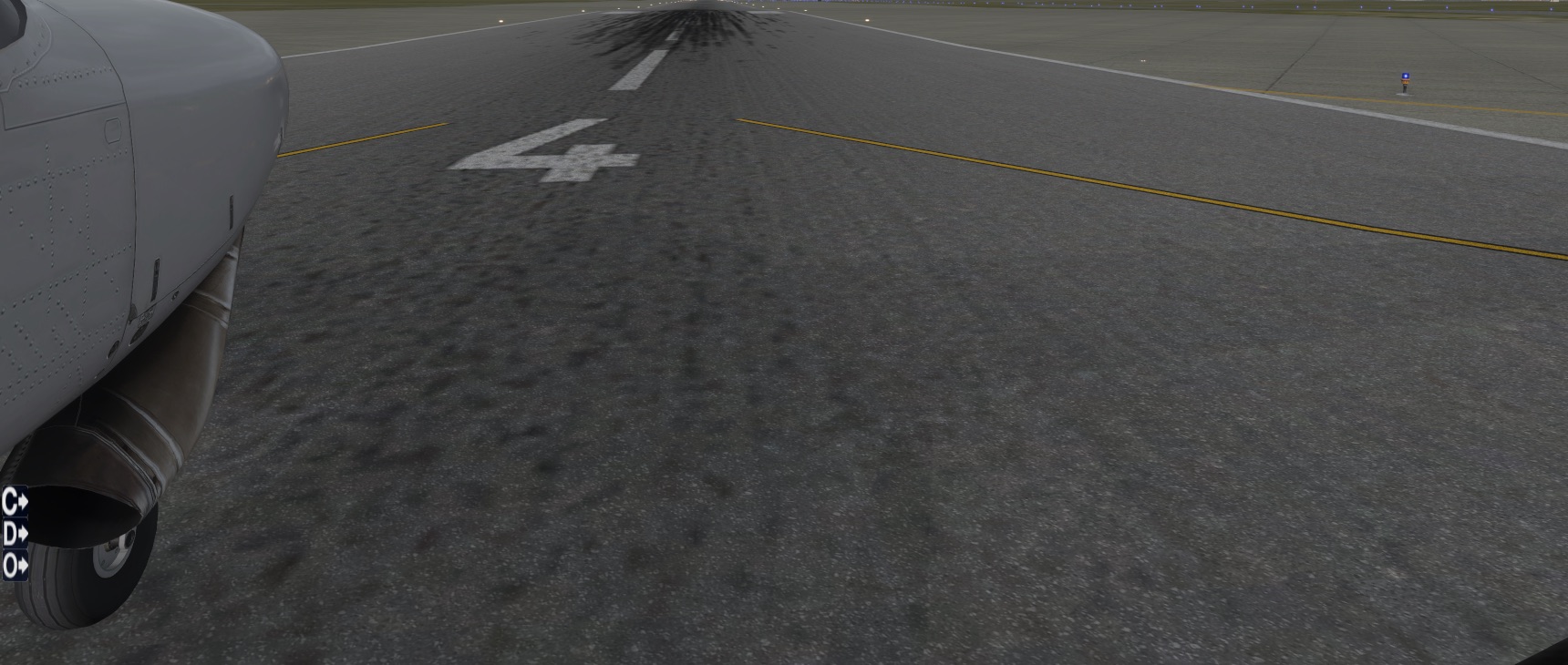}
\caption{Example input images to \taxinet, rendered in X-Plane. Left/right = clear/cloudy weather. Top/bottom = 12 pm / 4 pm.}
\label{fig:images}
\end{figure}

\taxinet{}'s neural network estimates the aircraft's position from a camera image; the camera is mounted on the right wing and faces forward.
Example images are shown in Fig.~\ref{fig:images}.
From such an image, the network estimates the \emph{cross-track error (CTE)}, the left-right offset of the plane from the centerline, and the \emph{heading error (HE)}, the angular offset of the plane from directly down the centerline.
These estimates are fed into a handwritten controller which outputs (the equivalent of) a steering angle for the plane.

The Boeing team provided the Berkeley team with an initial version of \taxinet{} without describing which images were used to train it.
In this way, the Berkeley team were not aware in advance of potential gaps in the training set and corresponding potential failure cases\footnote{After drawing conclusions from initial runs of all the experiments, the Berkeley team were informed of the training parameters and trained their own version of \taxinet{} locally, repeating the experiments. This was done in order to ensure that minor differences in the training/testing platforms at Boeing and Berkeley did not affect the results (which was in fact qualitatively the case). All numerical results and graphs use data from this second round of experiments.}.
For retraining experiments, the same sizes of training and validation sets were used as for the original model, as well as identical training hyperparameters.

The semantic feature space defined by our \scenic{} programs and searched by \verifai{} was 6-dimensional, made up of several parameters\footnote{We originally had additional parameters controlling the position and appearance of a tire mark superimposed on the runway (using a custom X-Plane plugin to do such rendering), but deleted the tire mark for simplicity after experiments showed its effect on \taxinet{} was negligible.}:
\begin{itemize}
\item the initial position and orientation of the aircraft (in 2D, on the runway);
\item the type of clouds, out of 6 discrete options ranging from clear to stormy;
\item the amount of rain, as a percentage;
\item the time of day.
\end{itemize}
Given values for these parameters from \verifai{}, the test protocol we used in all of our experiments was identical: we set up the initial condition described by the parameters, then simulated \taxinet{} controlling the plane for 30 seconds.

The main requirement for \taxinet{} provided by Boeing was that it should always track the centerline of the runway to within 1.5 m.
Encoded as an MTL formula for use with \verifai{}, this property is $\alwaysspec = \square (\text{CTE} \le 1.5)$.
For many of our experiments we created a greater variety of test scenarios by allowing the plane to start up to 8 m off of the centerline: in such cases we weakened the specification to $\evenspec = \lozenge_{[0, 10]} \square (\text{CTE} \le 1.5)$, which asserts that within 10 seconds the plane must reach within 1.5 m of the centerline and then stay there for the remainder of the simulation.

While both of these specifications are true/false properties, \verifai{} works with a continuous quantity $\rho$ called the \emph{robustness} of an MTL formula~\cite{mtl-robustness}.
Its crucial property is that when the formula is satisfied, $\rho \ge 0$, while if the formula is violated then $\rho \le 0$.
This means that $\rho$ provides a metric of \emph{how close} the system is to violating the property.
The exact definition of $\rho$ is not important here, but as an illustration, for $\alwaysspec$ it is equal to the greatest deviation beyond the allowed 1.5 m achieved over the whole simulation.

For additional experimental results, see Appendix~\ref{sec:appendix}.

\subsection{Falsification} \label{sec:falsification}

\begin{figure}[tb]
\centering
\includegraphics[width=0.8\textwidth]{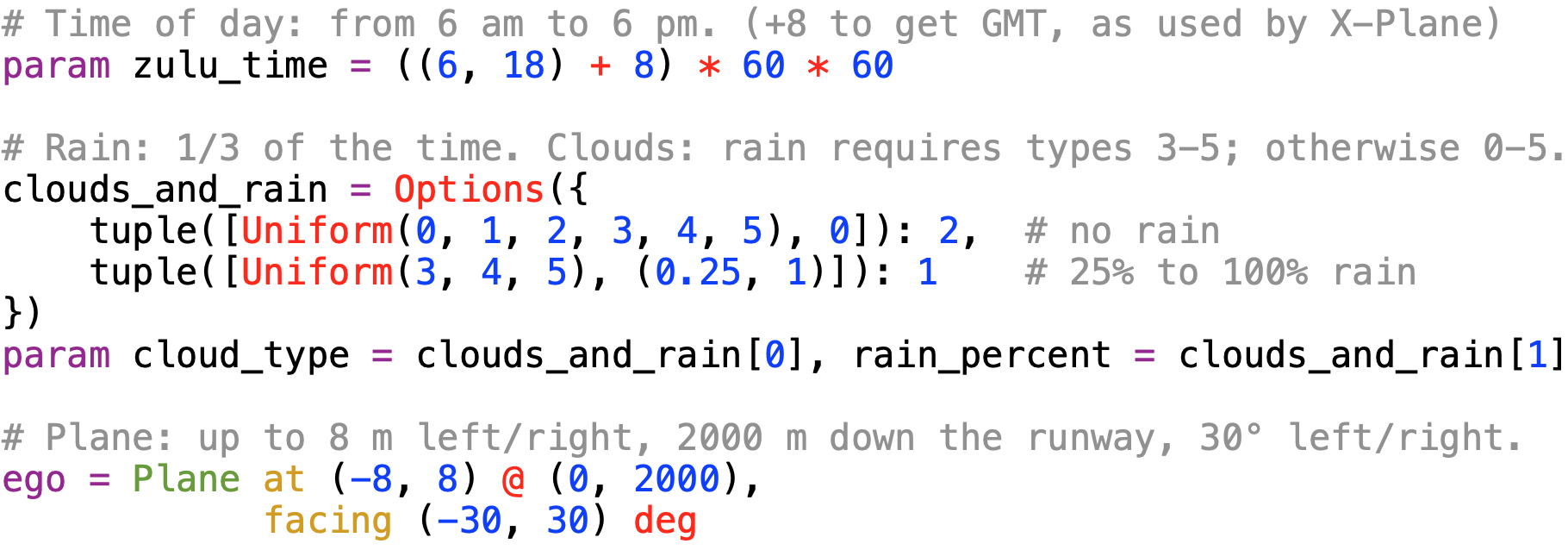}
\caption{Generic \scenic{} program \sfalsif{} used for falsification and retraining.}
\label{fig:rand-8m-30deg}
\end{figure}

In our first experiment, we searched for conditions in the nominal operating regime of \taxinet{} which cause it to violate \evenspec.
To do this, we wrote a \scenic{} program modeling that regime, shown in Fig.~\ref{fig:rand-8m-30deg}.
We first place a uniform distribution on time of day between 6 am and 6 pm local time (approximate daylight hours).
Next, we determine the weather.
Since only some of the cloud types are compatible with rain, we put a joint distribution on them: with probability $2/3$, there is no rain, and any cloud type is equally likely; otherwise, there is a uniform amount of rain between $25\%$ and $100\%$\footnote{The $25\%$ lower bound is because we observed that X-Plane seemed to only render rain at all when the rain fraction was around that value or higher.}, and we allow only cloud types consistent with rain.
Finally, we position the plane uniformly up to 8 m left or right of the centerline, up to 2000 meters down the runway, and up to $30^\circ$ off of the centerline.
These ranges ensured that (1) the plane began on the runway and stayed on it for the entire simulation when tracking succeeded, and (2) it was always possible to reach the centerline within 10 seconds and thereby satisfy $\evenspec$.

However, it was quite easy to find falsifying initial conditions within this scenario.
We simulated over 4,000 randomly-sampled runs, and found many counterexamples: in only 55.2\% of the runs did \taxinet{} satisfy $\evenspec$, and in 9.1\% of runs, the plane left the runway entirely.
This showed that \taxinet{}'s behavior was problematic, but did not explain \emph{why}.
To answer that question, we analyzed the data \verifai{} collected during falsification, as we explain next.

\subsection{Error Analysis and Debugging}

\verifai{} builds a table which stores for each run the point sampled from the abstract feature space and the resulting robustness value $\rho$ (see Sec.~\ref{sec:setup}) for the specification.
The table is compatible with the \emph{pandas} data science library~\cite{pandas}, making visualization easy.
While \verifai{} contains algorithms for automatic analysis of the table (e.g.~clustering and Principal Component Analysis), we do not use them here since the parameter space was low-dimensional enough to identify failure cases by direct visualization.

\begin{figure}[tbp]
\centering
\includegraphics[width=0.95\textwidth]{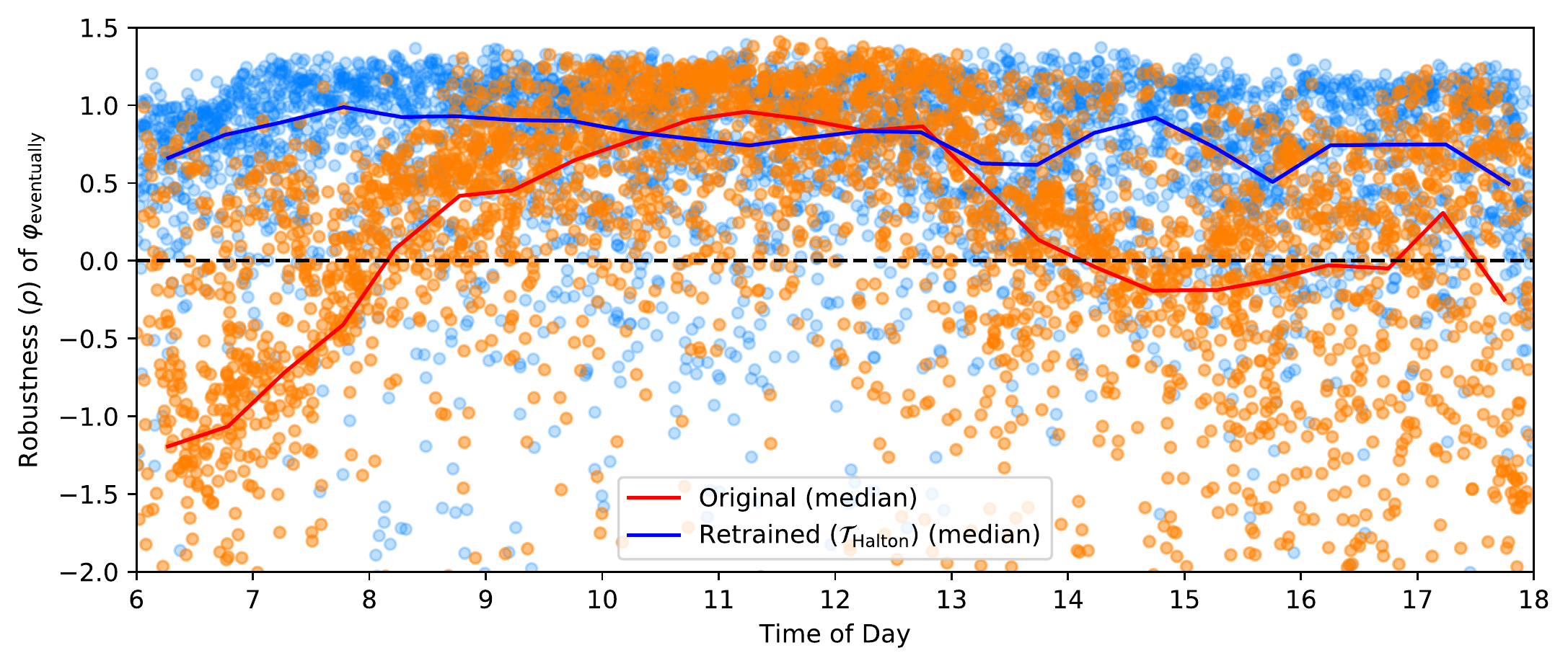}
\caption{Performance of \taxinet{} as a function of time of day, before and after retraining.}
\label{fig:time}
\end{figure}

\begin{figure}[p]
\centering
\includegraphics[width=0.9\textwidth]{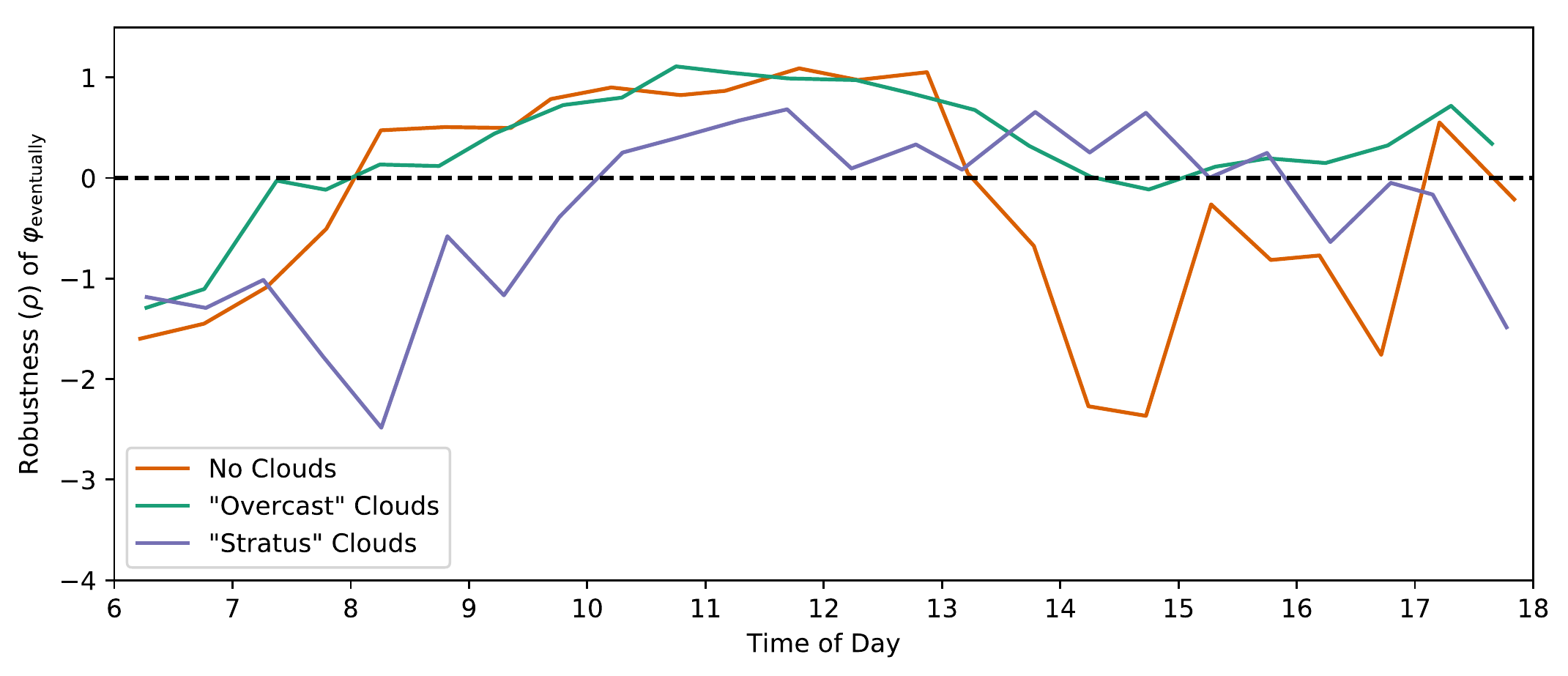}
\caption{Median \taxinet{} performance by time of day, for different cloud types. (For clarity, individual runs are not shown as dots in this figure.)}
\label{fig:clouds}
\end{figure}

\begin{figure}[p]
\centering
\includegraphics[width=0.9\textwidth]{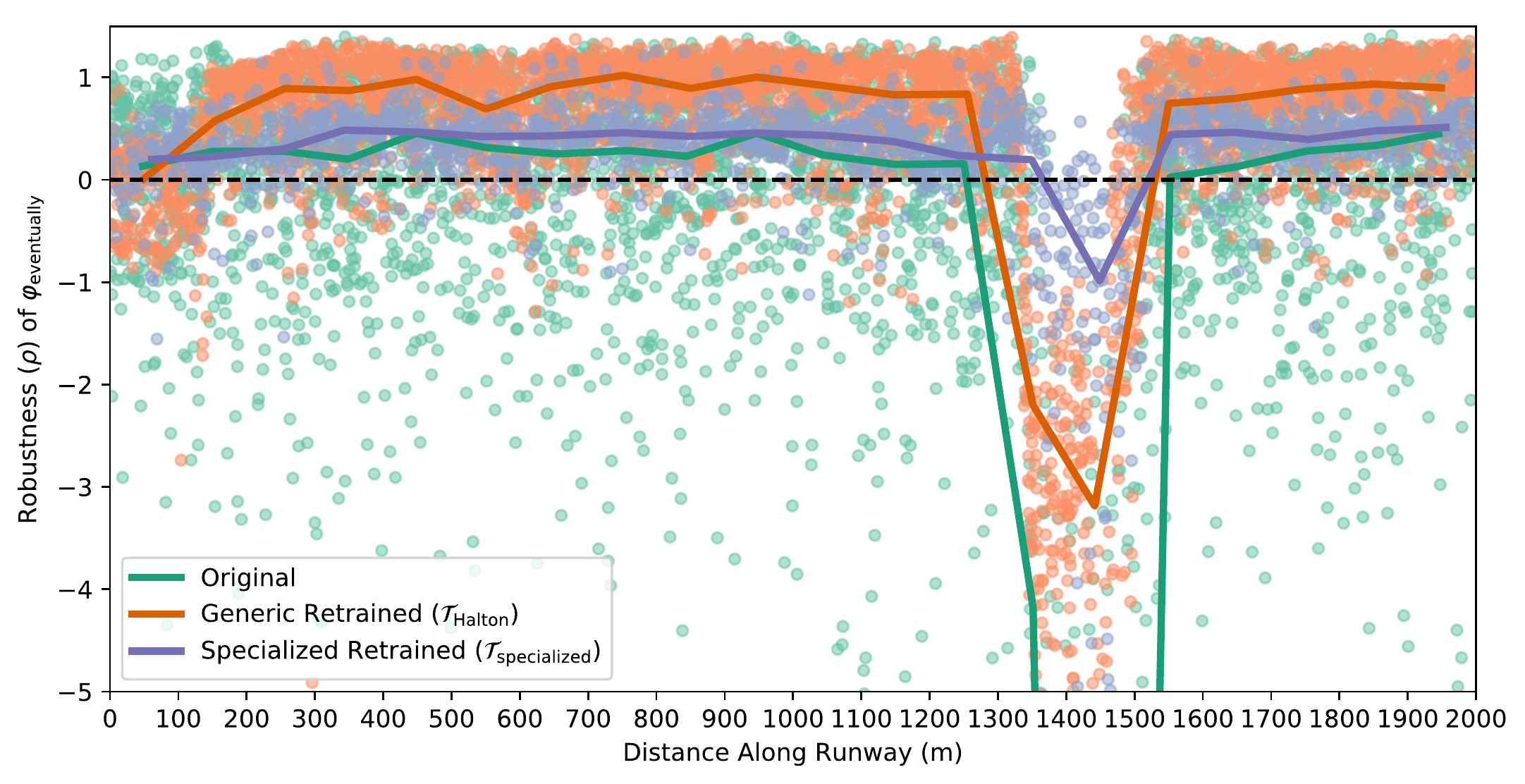}
\caption{\taxinet{} performance by distance along the runway. Solid lines are medians. The lowest median value for original \taxinet{} clipped by the bottom of the chart is $-32$.}
\label{fig:runway}
\end{figure}

We began by plotting \taxinet{}'s performance as a function of each of the parameters in our falsification scenario.
Several parameters had a large impact on performance:
\begin{itemize}
    \item \textbf{Time of day:} Figure~\ref{fig:time} plots $\rho$ vs. time of day, each orange dot representing a run during falsification; the red line is their median, using 30-minute bins (ignore the blue data for now).
    Note the strong time-dependence: for example, \taxinet{} works well in the late morning (almost all runs having $\rho > 0$ and so satisfying $\evenspec$) but consistently fails to track the centerline in the early morning.
    
    \item \textbf{Clouds:} Figure~\ref{fig:clouds} shows the median performance curves (as in Fig.~\ref{fig:time}) for 3 of X-Plane's cloud types: no clouds, moderate ``overcast'' clouds, and dark ``stratus'' clouds.
    Notice that at 8 am \taxinet{} performs much worse with stratus clouds than no clouds, while at 2 pm the situation is reversed.
    Performance also varies quite irregularly when there are no clouds --- we will analyze why this is the case shortly.
    
    \item \textbf{Distance along the runway:} The green data in Fig.~\ref{fig:runway} show performance as a function of how far down the runway the plane starts (ignore the orange/purple data for now).
    \taxinet{} behaves similarly along the whole length of the runway, except around 1350--1500 m, where it veers completely off of the runway ($\rho \approx -30$).
    Consulting the airport map, we find that another runway intersects the one we tested with at approximately 1450 m.
    Images from the simulations show that at this intersection, both the centerline and edge markings of our test runway are obscured.
\end{itemize}

\begin{figure}[tbp]
\centering
\includegraphics[width=0.9\textwidth]{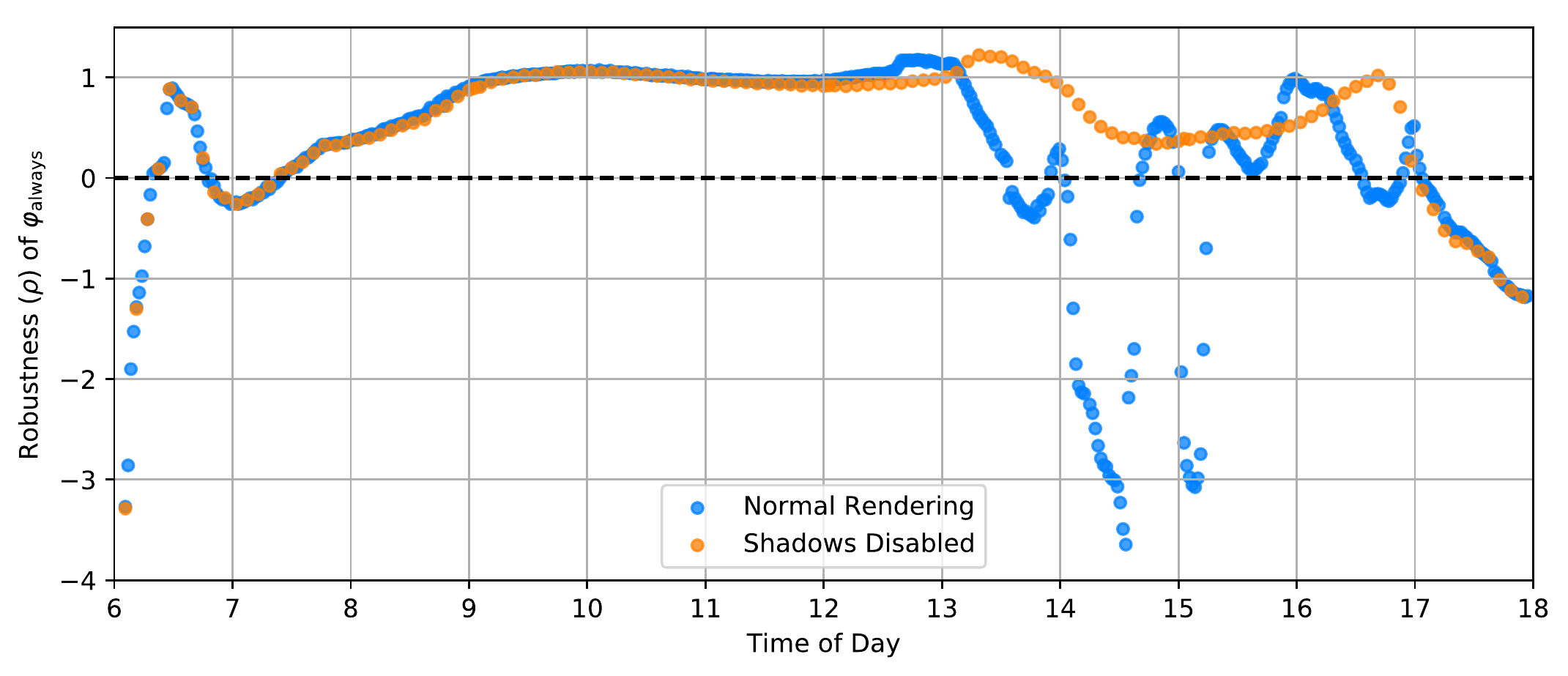}
\caption{\taxinet{} performance (with fixed plane position) by time of day, with and without shadows.}
\label{fig:shadows}
\end{figure}

These visualizations identify several problematic behaviors of \taxinet{}: consistently poor performance in the early morning, irregular performance at certain times depending on clouds, and an inability to handle runway intersections.
The first and last of these are easy to explain as being due to dim lighting and obscured runway markings.
The cloud issue is less clear, but \verifai{} can help us to debug it and identify the root cause.

Inspecting Fig.~\ref{fig:clouds} again, observe that performance at 2--3 pm with no clouds is poor.
This is surprising, since under these conditions the runway image is bright and clear; the brightness itself is not the problem, since \taxinet{} does very well at the brightest time, noon.
However, comparing images from a range of times, we noticed another difference: shortly after noon, the plane's shadow enters the frame, and moves across the image over the course of the afternoon.
Furthermore, the shadow is far less visible under cloudy conditions (see Fig.~\ref{fig:images}).
Thus, we hypothesized that \taxinet{} might be confused by the strong shadows appearing in the afternoon when there are no clouds.

To test this hypothesis, we wrote a new \scenic{} scenario with no clouds, varying only the time of day; we used \verifai{}'s Halton sampler~\cite{halton} to get an even spread of times with relatively few samples.
We then ran two experiments: one with our usual test protocol, and one where we disabled the rendering of shadows in X-Plane.
The results are shown in Fig.~\ref{fig:shadows}: as expected, in the normal run there are strong fluctuations in performance during the afternoon, as the shadow is moving across the image; with shadows disabled, the fluctuations disappear.
This confirms that shadows are a root cause of \taxinet{}'s irregular performance in the afternoon.

Figures~\ref{fig:time} and \ref{fig:runway} show that there are failures even at favorable times and runway positions.
We diagnosed several additional factors leading to such cases, such as starting at an extreme angle or further away from the centerline; see Appendix~\ref{sec:appendix} for details.

Finally, we can use \verifai{} for fault localization, identifying which part of the system is responsible for an undesired behavior.
\taxinet{}'s main components are the neural network used for perception and the steering controller: we can test which is in error by replacing the network with ground truth CTE and HE values and testing the counterexamples we found above again.
Doing this, we found that the system always satisfied $\evenspec$; therefore, all the failure cases were due to mispredictions by the neural network.
Next, we use \verifai{} to retrain the network and improve its predictions.

\subsection{Retraining}

The easiest approach to retraining using \verifai{} is simply to generate a new generic training set using the falsification scenario \sfalsif{} from Fig.~\ref{fig:rand-8m-30deg}, which deliberately includes a wide variety of different positions, lighting conditions, and so forth.
We sampled new configurations from the scenario, capturing a single image from each, to form new training and validation sets with the same sizes as for original \taxinet{}.
We used these to train a new version of \taxinet{}, \tngen{}, and evaluated it as in the previous section, obtaining much better overall performance: out of approximately 4,000 runs, 82\% satisfied $\evenspec$, and only 3.9\% left the runway (compared to 55\% and 9.1\% before).
A variant of \tngen{} using \verifai{}'s Halton sampler, \tnhalton{}, was even more robust, satisfying $\evenspec$ in 83\% of runs and leaving the runway in only 0.6\% (a $15\times$ improvement over the original model).
Furthermore, retraining successfully eliminated the undesired behaviors caused by time-of-day and cloud dependence: the blue data in Fig.~\ref{fig:time} shows the retrained model's performance is consistent across the entire day, and in fact this is the case for each cloud type individually.

However, this na\"ive retraining did not eliminate all failure cases: the orange data in Fig.~\ref{fig:runway} shows that \tnhalton{} still does not handle the runway intersection well.
To address this issue, we used a second approach to retraining: over-representing the failure cases of interest in the training set using a specialized \scenic{} scenario~\cite{scenic}.

We altered \sfalsif{} as shown in Fig.~\ref{fig:bump20}, increasing the probability of the plane starting 1200--1600 m along the runway, a range which brackets the intersection; we also emphasized the range 0--400 m, since Fig.~\ref{fig:runway} shows the model also has difficulty at the start of the runway.
We trained a specialized model \tnspec{} using training data from this scenario together with the validation set from \tngen.
The new model had even better overall performance than \tnhalton{}, with 86\% of runs satisfying $\evenspec$ and 0.5\% leaving the runway.
This is because performance near the intersection is significantly improved, as shown by the purple data in Fig.~\ref{fig:runway}; however, while the plane rarely leaves the runway completely, it still typically deviates several meters from the centerline.
Furthermore, performance is worse than \tngen{} and \tnhalton{} over the rest of the runway, suggesting that larger training sets might be necessary for further performance improvements.

\begin{wrapfigure}[12]{r}{0.42\textwidth}
\centering
\begin{bscenario}
rd = Options({
  (0, 400): 0.35,     # 0.2
  (400, 1200): 0.1,   # 0.4
  (1200, 1600): 0.5,  # 0.2
  (1600, 2000): 0.05  # 0.2
})
ego = Plane at (-8, 8) @ rd
\end{bscenario}
\caption{Position distribution emphasizing the runway beginning and intersection. Probabilities corresponding to the original scenario (Fig.~\ref{fig:rand-8m-30deg}) shown in comments.}
\label{fig:bump20}
\end{wrapfigure}
While in this case it was straightforward to write the \scenic{} program in Fig.~\ref{fig:bump20} by hand, we can also \emph{learn} such a program automatically: starting from $\sfalsif$ (Fig.~\ref{fig:rand-8m-30deg}), we use cross-entropy sampling to move the distribution towards failure cases.
Applying this procedure to \tngen{} for around 1200 runs, \verifai{} indeed converged to a distribution concentrated on failures.
For example, the distribution of distances along the runway gave $\sim$79\% probability to the range 1400--1600 m, 16\% to 1200--1400 m, and 5\% to 0--200, with all other distances getting only $\sim$1\% in total.
Referring back to Fig.~\ref{fig:runway}, we see that these ranges exactly pick out where \tnhalton{} (and \tngen{}) has the worst performance.

Finally, we also experimented with a third approach to retraining, namely augmenting the existing training and validation sets with additional data rather than generating completely new data as we did above.
The augmentation data can come from counterexamples found during falsification~\cite{cegda}, from a handwritten \scenic{} scenario, or from a failure scenario learned as we saw above.
However, we were not able to achieve better performance using such iterative retraining approaches than simply generating a larger training set from scratch, so we defer discussion of these experiments to the Appendix.

%% file: conclusion.tex
\section{Conclusion} \label{sec:conclusion}

In this paper, we demonstrated \verifai{} as an integrated toolchain useful throughout the design process for a realistic, industrial autonomous system.
We were able to find multiple failure cases, diagnose them, and in some cases fix them through retraining.
We interfaced \verifai{} to the X-Plane flight simulator, and extended the \scenic{} language with external parameters, allowing the combination of probabilistic programming and active sampling techniques.
These extensions are publicly available~\cite{verifai,scenic-www}.

While we were able to improve TaxiNet's rate of satisfying its specification from 55\% to 86\%, a 14\% failure rate is clearly not good enough for a safety-critical system (noting of course that TaxiNet is a simple prototype not intended for deployment).
In future work, we plan to explore a variety of ways we might further improve performance, including repeating our falsify-debug-retrain loop (which we only showed a single iteration of), increasing the size of the training set, and choosing a more complex neural network architecture.
We also plan to further automate error analysis, building on the clustering and other techniques already integrated into \verifai{}, and to incorporate white-box reasoning techniques to improve the efficiency of search.

%% file: more-experiments.tex
\section{Additional Experimental Results} \label{sec:appendix}

\subsection{Iterative Retraining Experiments} \label{sec:iterative-experiments}

Starting with the network from \tngen, we did 20 epochs of further training on the \tngen{} training and validation sets augmented by 20\% with one of three types of images:
\begin{itemize}
    \item more images from the generic \scenic{} scenario \sfalsif{} (Fig.~\ref{fig:rand-8m-30deg}), as a baseline;
    \item runway intersection images, obtained using a modified \sfalsif{} scenario restricted to 1350--1550 m down the runway;
    \item images sampled from the failure distribution learned by the cross-entropy sampler in Sec.~\ref{sec:falsification}.
\end{itemize}

As we would expect, the baseline setup had somewhat better performance than \tngen{} due to its larger training set, with 85\% of runs satisfying \evenspec{} (vs. 82\% for \tngen{}).
However, both other models actually had worse performance than the baseline, with the cross-entropy model satisfying the specification 82\% of the time and the model augmented with runway intersection images only 73\% of the time (the latter actually regressing from \tngen{}!).